\documentclass[10pt]{article} 
\usepackage[preprint]{tmlr}

\usepackage{amsmath,amsfonts,bm}

\def\eqref#1{equation~\ref{#1}}

\def\1{\bm{1}}

\DeclareMathAlphabet{\mathsfit}{\encodingdefault}{\sfdefault}{m}{sl}
\SetMathAlphabet{\mathsfit}{bold}{\encodingdefault}{\sfdefault}{bx}{n}

\usepackage{hyperref}
\usepackage{url}
\usepackage{graphicx} 
\usepackage{algorithm} 
\usepackage{algorithmic} 
\usepackage{listings} 
\usepackage{natbib} 
\usepackage{booktabs} 
\usepackage{makecell} 
\usepackage{amsmath}
\usepackage[section]{placeins} 

\title{Augmented Vision-Language Models: A Systematic Review}

\author{\name Anthony C Davis \email tony.davis@jhuapl.edu \\ 
      \addr Johns Hopkins University
      \AND
      \name Burhan Sadiq \email bsadiq1@jhu.edu \\
      \addr Johns Hopkins University
      \AND
      \name Tianmin Shu \email tianmin.shu@jhu.edu \\
      \addr Johns Hopkins University
      \AND
      \name Chien-Ming Huang \email chienming.huang@jhu.edu \\
      \addr Johns Hopkins University
      }

\def\month{MM}  
\def\year{2025} 
\def\openreview{\url{https://openreview.net/forum?id=XXXX}} 

\begin{document}

\maketitle

\begin{abstract}
Recent advances in visual-language machine learning models have demonstrated exceptional ability to use natural language and understand visual scenes by training on large, unstructured datasets. However, this training paradigm cannot produce interpretable explanations for its outputs, requires retraining to integrate new information, is highly resource-intensive, and struggles with certain forms of logical reasoning. One promising solution involves integrating neural networks with external symbolic information systems, forming neural symbolic systems that can enhance reasoning and memory abilities. These neural symbolic systems provide more interpretable explanations to their outputs and the capacity to assimilate new information without extensive retraining. Utilizing powerful pre-trained Vision-Language Models (VLMs) as the core neural component, augmented by external systems, offers a pragmatic approach to realizing the benefits of neural-symbolic integration. This systematic literature review aims to categorize techniques through which visual-language understanding can be improved by interacting with external symbolic information systems.
\end{abstract}

\section{Introduction}

\subsection{Motivation}
Vision-Language Models (VLMs) represent a significant leap forward in artificial intelligence (AI), showing remarkable abilities to interpret complex visual scenes and generate coherent natural language descriptions, powering advancements in tasks such as visual question answering (VQA) and image/video captioning \cite{radford2021learning, alayrac2022flamingo}. Trained on vast web-scale datasets, these models excel at mapping between visual inputs and textual concepts. However, this end-to-end training paradigm inherently limits their capabilities in several critical ways. VLMs often produce outputs without clear justifications, making them difficult to trust or debug \cite{rudin2022interpretable}. Integrating new factual knowledge or correcting errors typically requires resource-intensive retraining \cite{mitchell2021fast}. Furthermore, despite their semantic understanding, VLMs often struggle with tasks that require precise logical deduction, mathematical calculation (for example, accurate object counting), verifiable factual recall of entities within an image, and complex spatial reasoning \cite{mirzadeh2024gsmsymbolicunderstandinglimitationsmathematical, zhang2025visionlanguagemodelsrepresentspace}. These limitations hinder their deployment in high-stakes applications that require precision, reliability, and adaptability.

The concept of neural-symbolic systems offers a compelling theoretical direction to address these shortcomings by combining the perceptual strengths of neural networks (NN) with the precision and structured reasoning capabilities of symbolic systems \cite{besold2017neuralsymboliclearningreasoningsurvey}. The goal is to create hybrid systems that can perceive the world like neural networks but reason about it with logical rigor and access explicit knowledge like symbolic AI. However, operationalizing this vision presents challenges. Many traditional neural-symbolic techniques require tightly coupled integration or predefined symbolic structures, which can be rigid, difficult to train, and may impose strong human biases on how the neural and symbolic components ought to interact \cite{marcus2020decadeaistepsrobust}. For example:

\textit{Systems using fixed symbolic knowledge graphs}: Early approaches might involve converting a predefined knowledge graph into constraints or features for a neural network \cite{wang2017knowledge}. The structure of this graph and the integration method are designed manually, limiting flexibility if the underlying knowledge or required reasoning changes.

\textit{Rule injection methods}: Techniques like Knowledge Base Neural Networks (KBANN) \cite{towell1994knowledge} directly mapped predefined symbolic rules (e.g., Prolog) onto the initial structure and weights of a neural network. This tightly couples the network architecture to a specific human-authored rule set, making it rigid and difficult to adapt through further data-driven learning without disrupting the initial logic.

\textit{Hard-coded reasoning pipelines}: Some systems employ a pipeline where a neural module (e.g., object detector) extracts features or identifies objects, populating a symbolic representation (like a fact base or working memory). A separate, fixed symbolic reasoner, operating with predefined rules or logic (e.g., like the production rules in cognitive architectures such as Soar \cite{laird2022introductionsoar}), then processes these facts to draw conclusions or plan actions. The reasoning logic is hand-crafted, representing a strong bias about how perception and reasoning should be segregated and interact.

\textit{Logic Tensor Networks (LTNs) or similar frameworks}: While powerful, frameworks that translate First-Order Logic formulas directly into differentiable constraints for neural network training \cite{serafini2016logictensornetworksdeep} rely on humans specifying the exact logical axioms beforehand. This imposes the designer's assumptions about the domain's logic onto the learning process, which might be incomplete or subtly incorrect, and integrating complex logical constraints can make training optimization challenging.

Within this broader landscape, Augmented Vision-Language Models (AVLMs) emerge as a particularly pragmatic and promising implementation strategy. Instead of attempting a deep, complex fusion of neural and symbolic architectures from scratch, AVLMs leverage the power of existing, well-trained VLMs as a core component. Augmentation here refers to equipping these VLMs with the ability to interact with external, often symbolic, information sources or computational modules during their reasoning process. This approach offers distinct advantages:

\begin{itemize}
 \item \textit{Leverages existing strengths}: It builds upon the sophisticated visual and language understanding capabilities already present in state-of-the-art VLMs.

 \item \textit{Adaptability to diverse tasks}: By enabling interaction with various external resources (e.g., calculators, knowledge bases, application programming interfaces (APIs), specialized reasoners), a single VLM can be adapted to handle a wider range of tasks requiring specific computations or information retrieval \cite{qin2023toolllm, schick2023toolformerlanguagemodelsteach}. The system gains versatility by drawing upon the appropriate external capability for the problem at hand, rather than needing the core VLM to master every possible skill internally.

 \item \textit{Learnable integration}: Crucially, the VLM can often learn how and when to initiate these external interactions based on the input query and visual context, using standard machine learning techniques (e.g., fine-tuning). This data-driven approach to managing interactions is less rigid and potentially more adaptable than methods relying heavily on predefined symbolic rules, allowing the system to discover effective strategies for combining internal representations with external capabilities.

 \item \textit{Targeted weakness mitigation}: AVLMs can directly address specific VLM weaknesses through controlled external interactions. For instance, poor mathematical skills can be offset by invoking an external calculator; factual inaccuracies can be mitigated by querying a knowledge base; and notably, deficits in complex spatial reasoning  \cite{wang2024pictureworththousandwords, li2023evaluatingobjecthallucinationlarge, zhang2025visionlanguagemodelsrepresentspace} could potentially be addressed by interfacing with specialized geometric computation modules or accessing structured spatial information databases.
\end{itemize}

Therefore, augmenting VLMs with external interaction capabilities presents a powerful pathway towards building more robust, accurate, and versatile AI systems capable of complex visual reasoning. This approach moves beyond relying solely on the implicit knowledge encoded in model parameters, enabling VLMs to dynamically access and utilize external information and computational abilities. This systematic review will focus specifically on these Augmented Vision-Language Models (AVLMs), exploring the techniques used to bridge the gap between VLM representations and external symbolic resources.

\subsection{Augmented Vision-Language Models: Definition and Scope}

This review focuses on a specific, highly relevant subclass of augmented neural systems: Augmented Vision-Language Models. We define an Augmented Model as a system where external information or computational processes are actively integrated with a neural model's inference operations (before, during, or after its forward pass) to enhance its capabilities. We emphasize that this survey investigates only augmentations during inference and not training, to distinguish it from data augmentation techniques. This inference augmentation goes beyond mere prompting techniques (e.g., chain-of-thought \cite{wei2023chainofthoughtpromptingelicitsreasoning}), which primarily elicit latent reasoning abilities without incorporating external data or tools during inference.

A Vision-Language Model, for the purpose of this review, is defined as a machine learning model processing visual and/or textual input to generate natural language text outputs, encompassing tasks like VQA and captioning. Models performing tasks like object detection or classification without natural language generation are excluded.

Therefore, an AVLM is an VLM integrated with external symbolic information systems, APIs, databases, or other computational tools. A AVLM may involve modifications in VLM neural architecture, or it may involve pre- or post-processing of inputs or outputs of the VLM.
Regardless, these integrations aims to overcome the inherent limitations of standalone VLMs and represent a particularly compelling implementation of the augmented neural system concept. 

\subsection{Related Work and Knowledge Gap}
The quest to enhance neural models, particularly in the vision-language domain, by incorporating external knowledge or symbolic reasoning has spurred significant research, reflected in several existing surveys. Reviews on knowledge-enhanced multimodal learning \cite{Lymperaiou2022ASO, Zhao2023RetrievingMI, Wajid2023DeepLA} investigate integrating factual knowledge, often via knowledge graphs or retrieval augmentation, to improve tasks like captioning and VQA. Concurrently, surveys exploring neuro-symbolic approaches \cite{Aditya2019IntegratingKA, Senior2023GraphNN, Hitzler2022NeuroSymbolicSR, Khan2024ASO} examine the broader challenge of combining neural perception with symbolic reasoning, often focusing on graph neural networks, spatio-temporal logic, or commonsense knowledge integration for better scene understanding and reasoning. Specific areas like VQA have also been surveyed \cite{Jamshed2021NLPMV, Mostafa2020ComparativeSO}, tracing the evolution towards models capable of more complex reasoning, sometimes touching upon the need for external knowledge or structured representations.

While these surveys provide valuable context by covering knowledge integration, neuro-symbolic methods, and advances in VQA reasoning, they do not specifically offer a systematic review focused on the augmentation of VLMs through interaction with diverse external symbolic systems and tools. Existing reviews often focus on specific knowledge types (e.g., knowledge graphs) or broader neuro-symbolic theory. There is a knowledge gap in understanding the landscape of techniques specifically designed to connect modern VLMs with external symbolic resources in a flexible, often learned manner (i.e., tool use). Particularly, there is a lack of systematic analysis regarding how these augmentation techniques address core VLM challenges, such as their noted difficulties with precise spatial reasoning \cite{wang2024pictureworththousandwords, li2023evaluatingobjecthallucinationlarge, zhang2025visionlanguagemodelsrepresentspace}. Augmentation via external tools or information sources presents a potential pathway to compensate for such weaknesses by providing structured spatial information or enabling interactions with geometric reasoners, at least until VLM architectures intrinsically improve in these areas.

This systematic literature review aims to fill this gap by specifically categorizing and analyzing techniques where VLMs interact with external symbolic information systems or tools to enhance their vision-language understanding capabilities. We seek to provide a structured overview of how these augmentations are implemented, what types of external systems are used, and how they address the limitations of standard VLMs, with a particular interest in emerging tool-use paradigms and their application to challenging visual reasoning tasks.

\section{Overview: Three Stages of Vision-Language Fusion} 

The papers surveyed demonstrate a variety of techniques for augmenting vision-language models with external symbolic information systems. The selection of these studies is the result of a systematic literature search conducted according to the PRISMA guidelines, which involved querying academic databases with specific keywords and applying rigorous inclusion/exclusion criteria to identify relevant publications (see Appendix \ref{methodology}). This process ensures that the surveyed works specifically target inference-time augmentation and filter out approaches like pure prompting or training-time knowledge integration. To structure this diverse landscape, we categorize the surveyed approaches based on three key characteristics:

\begin{itemize}
\item \textit{When} the external interaction occurs relative to the VLM's processing pipeline. We distinguish between Early Fusion (integrating external data at the input stage, influencing initial representations), Middle Fusion (interfacing with external systems during the VLM's internal reasoning or generation steps), and Late Fusion (using the VLM's initial output to trigger external processing, validation, or refinement).

\item \textit{What} type of external information or computation is leveraged. This includes Retrieval (accessing pre-existing facts or knowledge from sources like knowledge graphs or text corpora) and Symbolic Computation (generating new information through logical deduction, program execution, or specialized computational tools), or a combination of both.

\item \textit{How} the fusion is specifically implemented, detailing the particular mechanisms used in each approach.
\end{itemize}

This review primarily organizes findings according to the temporal fusion stage (When), as this significantly impacts how external information influences the VLM. Within each temporal category (Early, Middle, Late), we further analyze the type of external interaction (What) and discuss notable implementation details (How). While some sophisticated methods may blend characteristics, this framework provides a structured lens for comparing the underlying principles, capabilities, and trade-offs of different augmentation techniques. The following sections elaborate on the findings for each category, referencing the detailed categorizations presented in the Appendix tables (Tables \ref{tab:early_fusion_survey_segmented_consistent_final} through \ref{tab:datasets_survey_segmented}).

\section{Early Fusion Methods}
Early fusion methods augment the VLM by incorporating external information directly at the input stage, before the core VLM begins its internal processing. This is often the conceptually simplest approach, treating external information as additional context and potentially requiring no VLM architecture changes. Its main advantage is implementation simplicity, offering a direct way to provide context. However, it faces challenges related to the relevance and noise of retrieved information. For example, some implementations use generated image captions as retrieved context which may introduce information loss. The choice between simple prompt augmentation and more structured retrieval encoding depends on the desired level of integration and complexity tolerance. These methods primarily fall into retrieval-based or, less commonly, symbolic computation-based categories, as detailed in Appendix Table \ref{tab:early_fusion_survey_segmented_consistent_final}.

\subsection{Retrieval-Based Early Fusion}
\begin{figure}[h]
\centering
\includegraphics[width=0.9\linewidth]{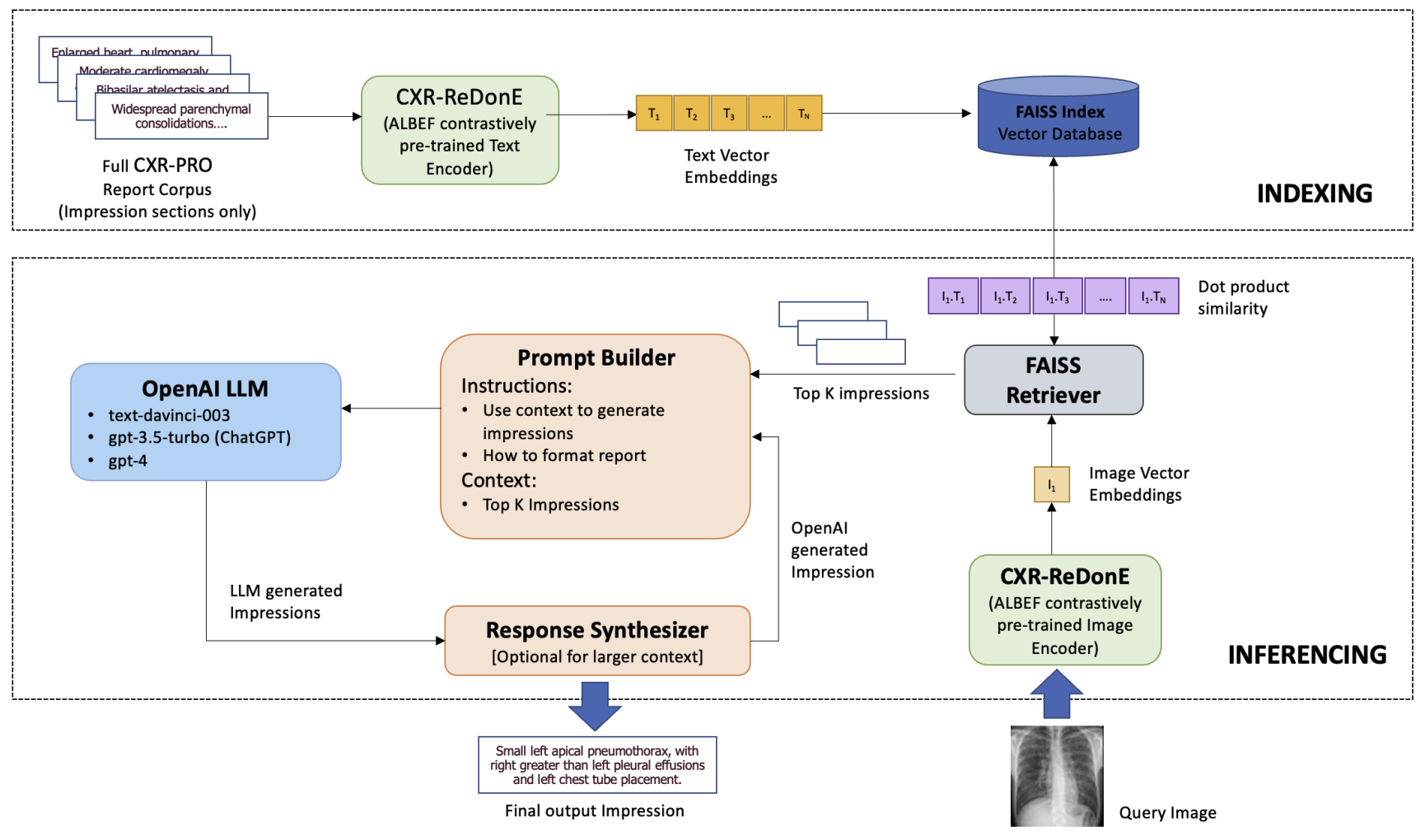}
\caption{Architecture for Retrieval Augmented Chest X-Ray Report Generation by \citet{Ranjit2023RetrievalAC}. Text embeddings from radiology impressions are indexed in a vector database. For an input X-ray image, its embedding, generated by a contrastively pretrained vision-language encoder (CXR-ReDonE), is used to retrieve the most similar text (impressions or sentences) from the database. This retrieved text then forms the context for a prompt, along with specific instructions, which is fed to an LLM (e.g., OpenAI GPT models) to generate the final radiology report impression. This process is illustrated for both indexing and inferencing stages.}
\label{fig:ranjit2023retrievalac}
\end{figure}

The most common early fusion strategy involves retrieving relevant information from external sources and providing it alongside the primary visual and textual inputs. A primary technique is \textbf{Prompt Augmentation}, where retrieved textual context is directly appended to the input prompt, exemplified by Retrieval Augmented Generation (RAG) \citep{lewis2021retrievalaugmentedgenerationknowledgeintensivenlp}. This retrieved text can originate from various sources. Text/Fact Retrieval draws information from text corpora or knowledge graphs (KGs), ranging from using pre-trained encoders like CLIP without further training \citep{Kan2023KnowledgeAwarePT, Ranjit2023RetrievalAC, Qu2024AlleviatingHI, Liu2024RARRA, Yan2024EchoSightAV, Xu2024ReverseIR, Khaliq2024RAGARYF, Xuan2024LEMMATL} to fine-tuning the retriever, possibly jointly with the VLM, for better relevance \citep{Iscen2023RetrievalEnhancedCV, Joshi2024RobustMM, Chen2022MuRAGMR, Gur2021CrossModalRA, Cui2024MOREMR, Zhu2023MultimodalNL, Hao2024KnowledgeCA}. See Figure \ref{fig:ranjit2023retrievalac} for an example of text retrieval using a pretrained vision-language encoder. Reranking retrieved results is often employed to enhance quality \citep{Qu2024AlleviatingHI, Liu2024RARRA, Wen2024MultimodalRF}. Retrieved KG triplets can also be formatted as text for the prompt \citep{Ravi2022VLCBERTVQ, Narasimhan2018StraightTT, Vickers2021InFE, Guo2022AUE, Wang2015ExplicitKR, Natu2023ExternalCK, Jhalani2024PrecisionEE, Barezi2024FindTG, Zhang2024GRACEGC, Zhang2023KnowledgeAwareCI, Wang2023DifferentiableOD, Ogawa2024PredictionOA, Chen2022LaKoKV, Kan2021ZeroShotSG, Gan2023OpenSetKV, Yang2019KnowledgeableSA}. An alternative form of prompt augmentation uses Image Caption Augmentation, where textual descriptions (captions, labels, Optical Character Recognition (OCR)) are first generated from the visual input, and this text is then used for retrieval or directly added to the prompt \citep{Gao2022TransformRetrieveGenerateNL, An2024KnowledgeAD, Li2018VisualQA, Fabian2023MultimodalFM, Sharifymoghaddam2024UniRAGUR, Ghosal2023LanguageGV, Khademi2023MMReasonerAM, Fu2023GenerateTS, Dey2021ExternalKA, Lin2022REVIVERV, Yu2019MultisourceMA, Liu2024CounterfactualVD, Mogadala2019MultiviewRL}. Some methods jointly train the caption generator and retriever \citep{Lin2022RetrievalAV, Garcia-Olano2021ImprovingAD, Salemi2023PreTrainingMD, Luo2021WeaklySupervisedVF, Vo2022NOCREKNO, Hao2024SelfBootstrappedVM, Gui2021KATAK, Chen2021KBVLPKB, Liang2021MariaAV, Lerner2023MultimodalIC}. While simplifying the problem to text-based retrieval, this approach risks information loss during captioning.

Instead of appending raw text, another approach uses \textbf{Retrieval Encoders} to encode the retrieved information (e.g., KG subgraphs, text passages) into separate embedding vectors. These embeddings then condition the VLM, often through attention mechanisms \citep{Yuan2023RAMMRB, Weng2024LearningTC, Chen2022KRITKI, Salemi2023ASD}, Long Short Term Memory models (LSTMs) \citep{Wu2016ImageCA}, or memory modules \citep{Hu2022RevealRV}. This allows for a more structured integration of knowledge. Specifically, KG subgraphs can be encoded using Graph Neural Networks (GNNs) \citep{Li2020BoostingVQ, Rao2023RetrievalbasedKA, Zhang2020RichVK, Lee2021VisionLanguageKnowledgeCF, Li2022SupportingVM, Lin2023TowardsMA, Torino2020SemanticsAwareVA, Wang2022KnowledgeEnhancedVQ, Qu2020KSFSTVC, Padhi2024ImprovingCC, Shevchenko2021ReasoningOV, Gardres2020ConceptBertCR, Jing2023MultisourceSG, Mondal2024KAMCoTKA, Lee2024MultimodalRW} or fused with scene graphs \citep{Chen2021ZeroshotVQ, Ziaeefard2020TowardsKV, Yu2020CrossmodalKR, Zhu2020MuckoMC, Hussain2022MultimodalKR, Li2022DynamicKM, Ye2021BreakingSB}. Multimodal KGs can also provide richer representations \citep{Jiang2023KnowledgeBasedVQ}.

\subsection{Symbolic Computation Early Fusion}
Integrating the results of symbolic computations at the input stage is rare in the surveyed literature. The primary example identified \citep{Potapov2019CognitiveMN} involves transforming the visual input into a symbolic scene graph. This structured representation, potentially processed by an external symbolic reasoning engine like OpenCog, serves as input or conditioning for the VLM. This approach explicitly introduces symbolic structure early on but depends heavily on robust perception-to-symbol conversion modules.

\section{Middle Fusion Methods}
Middle fusion techniques integrate external information or symbolic computation \textit{during} the VLM's forward pass, allowing interaction with the model's intermediate representations. This enables more dynamic and potentially iterative integration compared to early fusion, where external data influences internal processing, reasoning steps, or feature refinement. By allowing external information and symbolic processes to interact with the VLM's internal state, these methods enable context-aware reasoning and iterative refinement. This often involves more complex architectures and training but holds promise for leveraging both neural pattern recognition and symbolic manipulation more effectively. The rise of tool use and agent-based frameworks within this category points towards VLMs acting less as monolithic predictors and more as components in larger reasoning systems, echoing paradigms like Kahneman's System 1 (neural intuition) and System 2 (deliberate symbolic reasoning) \citep{kahneman_thinking_2011}. These methods, categorized in Appendix Table \ref{tab:middle_fusion_survey_segmented}, often involve feedback loops or specialized modules operating alongside main VLM components.

\subsection{Retrieval-Based Middle Fusion}
These methods retrieve external information based on intermediate VLM states and fuse it back into the ongoing computation. One approach is \textbf{Dense Retrieval}, which uses dense vector similarity between intermediate VLM representations and a knowledge corpus to find relevant information (often images or text) that is then fused back into the model's layers, typically via attention \citep{Wang2022VisuallyAugmentedLM, Lin2023FinegrainedLM, Jia2023KAFARI}. Another major approach leverages \textbf{Graph-Based Retrieval}, primarily using KGs. This includes methods where intermediate visual or textual features trigger \textbf{KG Querying}; the retrieved subgraphs or facts are processed (often with GNNs) and fused with VLM representations \citep{Li2017IncorporatingEK, Li2023GraphAdapterTV, Zheng2021KnowledgeBG, Su2018LearningVK, 2018OutOT, Zhang2023KnowledgeenhancedVP, Singh2019FromST, Jiang2020UnderstandingCI, Yu2023KnowledgeAwareGR, Du2022CALMCK, Li2024HKFNetFE, Yin2023MultiClueRW, Ma2022MultiSourceKR, Zhu2020ConfigurableGR, Cao2019ExplainableHV, Li2022LearningTR, ZhengSageAM}, sometimes after extracting visual subgraphs first \citep{Wei2022EnhanceUA, Narayanan2021VQAAA}. Other graph-based methods use \textbf{Similarity Measures} between internal VLM representations and KG elements to guide reasoning or weighting, rather than directly injecting KG structure \citep{Wu2024ResolvingZA, Chae2022UncertaintybasedVQ, Li2019SemanticCN, Xian2023MultimodalKT, Marino2020KRISPII}. A significant group focuses on \textbf{Concept/Scene Graph Fusion}, explicitly combining internally generated scene graphs with external concept graphs (e.g., from ConceptNet \citep{speer2018conceptnet55openmultilingual}), often using GNNs on the combined graphs \citep{Yang2023VisualQA, Wang2022VQAGNNRW, Khan2022NeuSIRENI, Khan2022ExpressiveSG, Zhu2022FromST, Wen2021MultiLevelKI, Song2023HowTU, Li2022IncorporatingEK, Zhang2021ExplicitKI, Dong2024ModalityAwareIW, Gao2023RoomObjectEP, Zhang2022ReasoningWM, Xu2021ImagineRA, Li2024IIUII, Hou2020JointCA, Gu2019SceneGG}. More complex structures like \textbf{Multimodal KGs} (MMKGs) \citep{Xi2024KnowledgeGE, Shi2022ImprovingZP, Santiesteban2024ImprovedIC, Ouyang2024ModaladaptiveKG, Liu2021FactbasedVQ} or \textbf{Hypergraphs} \citep{Heo2022HypergraphTW, Wang2024HyperMRHH} are also integrated using specialized graph networks. Finally, \textbf{Reinforcement Learning} (RL) can be used to learn policies for querying or integrating external knowledge based on the current state \citep{Bougie2018CombiningDR}.

\subsection{Symbolic Computation Middle Fusion}
\begin{figure}[h!]
    \centering
    \includegraphics[width=0.5\textwidth]{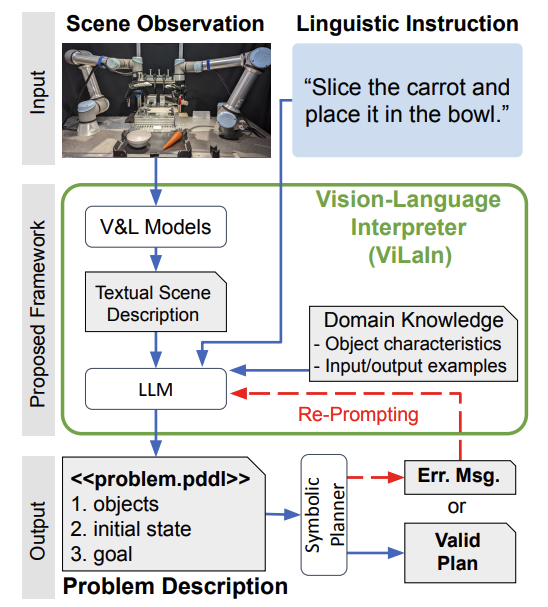}
    \caption{Overview of the ViLaIn approach for VLM planning of robotic actions \citet{Shirai2023VisionLanguageIF}. The vision-language interpreter (ViLaIn) generates a problem description from a linguistic instruction and scene observation. The symbolic planner finds an optimal plan from the generated problem description.}
    \label{fig:shirai2023}
\end{figure}

These methods incorporate symbolic reasoning, calculations, or tool use within the VLM's processing pipeline. One key technique is \textbf{Program Synthesis}, where the VLM generates intermediate programs (e.g., functional programs, Python code) operating on symbolic input representations or querying external tools; the execution result influences subsequent VLM processing \citep{Zhang2022QueryAA, Zhang2023TowardMD, Hu2023VisualPD}, \citep[see Figure~\ref{fig:shirai2023}]{Shirai2023VisionLanguageIF}, \citep{Zhang2023GroundingCT, Li2021CalibratingCA, Mishra2024LearningRF, Xue2024IntegratingNR}. Another approach involves integrating \textbf{Symbolic Logic Engines}, translating intermediate VLM representations into facts or queries processed by engines like differentiable first-order logic \citep{Amizadeh2020NeuroSymbolicVR}, Answer Set Programming (ASP) \citep{Riley2019IntegratingNL, Mitchener2021DetectUA}, Description Logic \citep{Tsatsou2021TowardsUK}, planning domain definition languages (PDDL) \citep{Zhang2022DANLIDA, Zhang2023SEAGULLAE}, temporal logic \citep{Choi2024TowardsNV}, specialized neurosymbolic languages like Scallop \citep{Li2023ScallopAL, Huang2021ScallopFP}, or embedding propositional logic operations \citep{Li2023AND}. \textbf{Vector Symbolic Architectures (VSAs)} represent symbols and perform operations using high-dimensional vectors within the neural architecture \citep{Montone2017HyperdimensionalCF, Kovalev2021VectorSM}. Some methods perform \textbf{Symbolic Graph Operations} directly on graph representations (scene graphs, KGs) during processing, like guided walks or routing \citep{Li2022InnerKI, Liang2020LRTAAT, Wu2023SymbolLLMLL, Zhao2015AQF, Yang2020ObjectCentricDO, Zhang2023InterpretableDO, Hudson2019LearningBA, Cao2021LinguisticallyRC}. Increasingly popular is \textbf{Tool Use}, where the VLM dynamically calls external tools (calculators, APIs, vision algorithms, drawing tools) based on its intermediate state, integrating the tool's output \citep{Hu2024VisualSS, Fan2024VideoAgentAM, Liu2023LLaVAPlusLT, Hu2023AVISAV, Wu2024AvaTaROL}. Lastly, \textbf{Self Play} involves using the VLM within a simulated environment where it interacts, uses tools (potentially itself), and learns from feedback \citep{Misiunas2024VQATS}.

\subsection{Combined Retrieval and Symbolic Computation Middle Fusion}
These advanced methods integrate both retrieval and symbolic computation during the forward pass. Many employ \textbf{Agent} architectures where the VLM acts as a controller, deciding when to retrieve information and when to use symbolic tools (including sub-agents or code execution) to achieve a goal \citep{Niu2024ScreenAgentAV, Castrejon2024HAMMRHM, Lu2023ChameleonPC, Hsieh2023ToolDE, Xu2024RSAgentAR, Yang2024DoraemonGPTTU}. \textbf{Other Approaches} combine retrieval (e.g., from ontologies, KGs) with symbolic reasoning (e.g., probabilistic logic, program synthesis, graph walks, concept binding) in bespoke ways for specific tasks like embodied QA, riddle solving, or rumor detection \citep{Besbes2015AnOV, Aditya2016AnsweringIR, Aditya2017ExplainableIU, Aditya2016DeepIUA, Tan2021KnowledgeBasedEQ, Liu2023KnowledgeEnhancedHI, Stammer2024NeuralCB, Vatashsky2018UnderstandCA, Gao2023ASR, Gao2024ExplainableVQ}.

\section{Late Fusion Methods}
Late fusion methods apply external information retrieval or symbolic computation \textit{after} the VLM has generated an initial output. This external step typically serves to validate, refine, explain, or augment the VLM's output using structured knowledge or precise tools. Late fusion provides a powerful mechanism for verification, refinement, and explanation by applying structured knowledge or precise computations to the VLM's generated output. It leverages the VLM's ability to produce a plausible initial response, which then guides a more targeted external process. This approach is particularly well-suited for enhancing reliability and interpretability, as symbolic steps can act as explicit checks or provide traceable reasoning paths. The main dependency is the quality of the initial VLM output; if it is too vague or incorrect, the subsequent external process may be misguided. These techniques are cataloged in Appendix Table \ref{tab:late_fusion_survey_segmented}.

\subsection{Retrieval-Based Late Fusion}
Here, the VLM's output triggers a targeted retrieval query. In \textbf{Dense Retrieval}, the initial VLM output (e.g., answer, rationale) queries a dense retrieval system. The retrieved information (text, facts) is then used to refine the output or provide supporting evidence \citep{Song2022AnsweringKV, Song2022EfficientAS, Shi2024QRCLIPIE}. Alternatively, using \textbf{Knowledge Graph Retrieval}, the VLM's output (e.g., generated caption, predicted relationships) queries a KG. Retrieved facts or subgraphs refine the output, for instance, by adjusting probabilities or improving relationship predictions \citep{Gao2022CrossModalOD, Huang2020BoostIC, Xiao2022VisualRD}.

\subsection{Symbolic Computation Late Fusion}
This involves applying symbolic tools or logic engines to the VLM's output. \textbf{Program Synthesis} generates programs based on the VLM's output for analysis, validation, or transformation. Examples include generating Python code to verify VQA answers via vision APIs \citep{Suris2023ViperGPTVI, Subramanian2023ModularVQ, Gupta2022VisualPC}, treating symbolic programs as latent variables \citep{Vedantam2019ProbabilisticNM}, or generating Structured Query Language (SQL) queries from the output \citep{Bhaisaheb2023ProgramSF}. The influential Neural-Symbolic VQA (NS-VQA) approach \citep{Yi2018NeuralSymbolicVD}, executing programs on scene representations post-prediction, is often adapted. \textbf{Symbolic Engines} feed the VLM's output (or derived symbolic representations) into formal logic engines (e.g., Prolog, ASP, Probabilistic Soft Logic) for consistency checking, inference, or validation \citep{Sethuraman2021VisualQA, Aditya2018SpatialKD, Eiter2022ANA, Eiter2021ACI, Cunnington2024TheRO}, or use PDDL for planning \citep{Xu2022SGLSG}. \textbf{Tool Use} involves calling external tools or APIs based on the VLM's output for specialized functions, verification, or generating structured data \citep{Yuan2023CRAFTCL, Cesista2024RetrievalAS, Cesista2024MultimodalSG, Zhang2023GraphToolFormerTE}. \textbf{Symbolic Graph Operations} perform manipulations on graph representations derived from the VLM's output, such as reasoning over action chains or graph traversals \citep{Li2023MultimodalAC, Zhan2021VisualQA, Saqur2020MultimodalGN, Johnston2023NSILNV}. \textbf{Other Approaches} include applying symbolic solvers to latent representations \citep{Singh2018ABA}, using VLM output confidence to trigger human interaction or further symbolic checks \citep{Bao2023ConfidencebasedIN}, or updating conversational memory based on the response \citep{Verheyen2023NeuroSymbolicPS}.

\subsection{Combined Retrieval and Symbolic Computation Late Fusion}
\begin{figure}[htbp]
    \centering
    \includegraphics[width=0.5\textwidth]{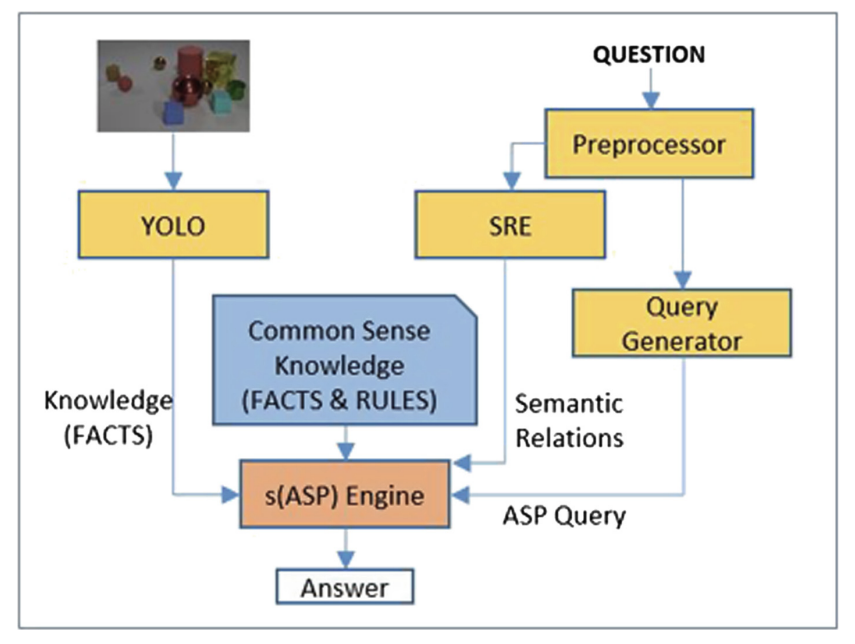}
    \caption{The architecture of the AQuA framework \citet{Basu2020AQuAAV}.
    It consists of five main modules: (i) YOLO for object detection and feature extraction,
    (ii) a Preprocessor for the natural language question, (iii) a Semantic Relation Extractor (SRE),
    (iv) a Query Generator based on semantic analysis, and (v) a retrieval based Commonsense Knowledge module
    leveraging. The system utilizes an ASP engine for symbolic reasoning.}
    \label{fig:aqua_architecture}
\end{figure}

These methods combine both retrieval and symbolic computation after the initial VLM output.
Typically, the VLM output is parsed into a logical form, relevant domain knowledge (facts or programs) is retrieved,
and a symbolic reasoner (e.g., probabilistic logic, ASP) derives the final answer \citep{Sachan2020TowardsLA, Basu2020AQuAAV}. The AQuA framework \citep{Basu2020AQuAAV} is depicted in Figure \ref{fig:aqua_architecture}.

\section{Datasets for Augmented VLMs}
The development and evaluation of augmented VLMs rely on suitable datasets, detailed in Appendix Table \ref{tab:datasets_survey_segmented}. These datasets often target capabilities where standard VLMs struggle, such as complex reasoning, external knowledge dependency, or precise spatial understanding. For \textbf{Spatial Reasoning}, datasets like Compositional Language and Elementary Visual Reasoning (CLEVR) \citep{johnson2016clevrdiagnosticdatasetcompositional} and its variants (CLEVRER \citep{yi2020clevrercollisioneventsvideo}, SuperCLEVR \citep{li2023superclevrvirtualbenchmarkdiagnose}, CLEVR Partially Observable Constraints (CLEVR-POC) \citep{Abraham2024CLEVRPOCRV}, SuperCLEVR-Physics \citep{Wang2024Compositional4D}) use controlled synthetic environments, while scene graph datasets like Visual Genome \citep{krishna2016visualgenomeconnectinglanguage} and Visual Commonsense Discovery (VCD) \citep{Shen2024VCDKB} offer real-world complexity. \textbf{Knowledge-Based VQA (KB-VQA)} datasets necessitate external knowledge; examples include VQA \citep{agrawal2016vqavisualquestionanswering}, Face-Based VQA (FVQA) \citep{Wang2016FVQAFV, Lin2023FVQA2I}, Knowledge-Aware VQA (KVQA) \citep{Shah2019KVQAKV}, Outside Knowledge VQA (OK-VQA) \citep{Marino2019OKVQAAV, Reichman2023OutsideKV}, Synthetic Knowledge VQA (SK-VQA) \citep{Su2024SKVQASK}, Encyclopedic VQA \citep{Mensink2023EncyclopedicVV}, Select, Substitute, and Search VQA (S3VQA) \citep{Jain2021SelectSS}, Knowledge Routed VQA \citep{Cao2020KnowledgeRoutedVQ}, Intensive-Neural-Knowledge (INK) \citep{Sung2022INKI}, and domain-specific ones like IndiFoodVQA \citep{Agarwal2024IndiFoodVQAAV}. Entity-specific datasets like SnapNTell \citep{Qiu2024SnapNTellEE} and ViQuAE \citep{Lerner2022ViQuAEAD} focus on entity-knowledge retrieval. \textbf{Reasoning-Based VQA} datasets test complex inference, including common sense or multi-step logic (e.g., A-OKVQA \citep{Schwenk2022AOKVQAAB}, CRIC \citep{Gao2019CRICAV}, VCR \citep{Zellers2018FromRT}, High Order Visual Question Reasoning (HVQR) \citep{Cao2019ExplainableHV}, visual riddles \citep{BittonGuetta2024VisualRA}). Some datasets require \textbf{Combined Knowledge and Spatial Reasoning}, such as InfoSeek \citep{Chen2023CanPV}, Situated Open-World Commonsense Reasoning (SOK-Bench) \citep{Wang2024SOKBenchAS}, and WikiTiLo \citep{Zhang2023CanVM}. Benchmarks for VLM-based \textbf{Agents} evaluate task completion via interaction with environments like the web, GUIs, or simulators (e.g., ScreenAgent \citep{Niu2024ScreenAgentAV}, WebArena/VisualWebArena \citep{zhou2023webarena, koh2024visualwebarenaevaluatingmultimodalagents}, Spider2 \citep{Cao2024Spider2VHF}). \textbf{Domain-Specific} datasets are crucial for applications like robotics \citep{Gao2023PhysicallyGV}, art explanation \citep{Hayashi2024TowardsAE}, fake news detection \citep{Jin2024FakeND}, and medical VQA \citep{Hu2023ExpertKI}. The trend is towards datasets demanding deeper reasoning, integration of diverse knowledge, and evaluation beyond accuracy to include interpretability and interaction, reflecting the maturing goals of AVLM research.

\section{Discussion}

The studies reviewed in this paper underscore the growing importance of incorporating external symbolic information into vision-language models across different fusion paradigms. Here we discuss key observations, challenges, and potential directions for future research stemming from these findings.

\subsection{Increasing Complexity in Integration Paradigms}
The review highlights a spectrum of integration strategies, ranging from early fusion (where external information is fed in parallel with raw visual or textual inputs) to middle fusion (symbolic or retrieval-based information is queried by and fed back into the VLM) and late fusion (where the VLM's output is used as input to an external information tool). As we move from early to late fusion, there is a noticeable increase in the sophistication of the symbolic or retrieval mechanisms. Early fusion methods often rely on prompt augmentation or direct inclusion of retrieved data, whereas middle and late fusion approaches rely on the VLM to interact with and query the symbolic/information system, effectively defining a new task that requires its own finetuning. These evolving paradigms reflect an attempt to balance practical system design with the quest for more interpretable and verifiable predictions. As of the time of writing, many commercially available AI products use only early fusion retrieval techniques, but as users demand increasingly more intelligent systems, more products will likely use middle fusion techniques to more effectively integrate VLMs with other information systems.

\subsection{Benefits of Neural-Symbolic Systems}
A recurring theme in the surveyed methods is the complementary strengths of neural and symbolic components. Neural networks excel at pattern recognition, approximate reasoning, and natural language understanding, while symbolic engines and structured knowledge bases provide exact memory, explicit logical constraints, and the ability to integrate new facts without retraining. Late fusion techniques offer the distinct capability of returning exact facts or conclusions, since the VLM is used like a multimodal query parsing engine. Across many tasks (particularly knowledge-intensive or reasoning-heavy tasks such as knowledge-based VQA, object-relation analysis, and robotics planning) hybrid architectures consistently outperform purely neural methods of similar computational requirements. Moreover, the explicitness of symbolic tools can boost interpretability, allowing users to trace how an answer was derived and why it is correct or incorrect.

\subsection{Trade-Offs in Complexity and Computation}
Despite the clear benefits of neural-symbolic approaches, engineering complexity emerges as a central challenge. Designing, maintaining, and updating knowledge graphs, database schemas, or symbolic modules for different tasks is non-trivial. Likewise, incorporating external tool use (such as specialized retrieval APIs, symbolic planners, or code interpreters) can significantly increase system complexity and inference latency, though some studies have shown that high quality documentation can provide enough information for well trained VLMs to use the tools without retraining. Regardless, in real-world applications, these trade-offs become critical: while early fusion methods may be relatively straightforward to implement, they can introduce noise through irrelevant retrieved information. Conversely, advanced middle or late fusion methods can be more precise but demand intricate coordination between neural and symbolic components. The computational overhead can also be substantial, especially for iterative or multi-step retrieval and symbolic reasoning.

\subsection{Benchmarks and Evaluation Gaps}
A variety of benchmarks exist, ranging from spatial reasoning datasets (CLEVR variants) to knowledge-based VQA sets (OK-VQA, FVQA, KVQA) and domain-specific tasks (medical imaging, robotics, computer control), yet the field still lacks a unifying evaluation framework. Although some datasets provide rationales or require multi-hop reasoning, few systematically evaluate both the correctness and interpretability of neural-symbolic systems in a consistent manner. Additionally, many benchmarks remain isolated, reflecting only a single domain or knowledge source. Future efforts should standardize evaluation protocols that measure not only accuracy but also interpretability, reasoning transparency, and efficiency.

\subsection{Potential for Iterative, Interactive, and Embodied Systems}
Many middle fusion techniques showcase the promise of iterative refinement and interactive loops, wherein the model queries external tools or knowledge bases multiple times, updating its internal representations at each step. This iterative paradigm reflects a “System 2” style of reasoning \cite{kahneman_thinking_2011}, complementing the quick “System 1” pattern matching that neural networks already excel at. Moreover, the emergence of agent-based systems and self-play frameworks illustrates an exciting trend where VLMs are empowered to act (through tool use, code generation, or robot control) and then revise or validate their own outputs based on feedback from the environment. In future research, a closer coupling of these agent-based approaches with robust knowledge resources and symbolic planning engines could yield more general and context-aware AI systems.

\subsection{Tool Use as a Unifying Abstraction for Augmentation}
Rather than seeking a single "best" method for neuro-symbolic integration, the concept of \textbf{tool use} emerges as a powerful and flexible abstraction layer for designing AVLMs \citep{qin2023toolllm, schick2023toolformerlanguagemodelsteach}. This perspective treats diverse external capabilities—whether retrieving from knowledge graphs \citep{Zhang2023GraphToolFormerTE}, executing code \citep{Suris2023ViperGPTVI}, performing calculations, querying databases, calling specialized perception modules like object detectors \citep{Gupta2022VisualPC}, or invoking complex symbolic reasoners \citep{Yang2024DoraemonGPTTU} as distinct tools accessible via a standardized interface. The literature also refers to this as function calling \citep{patil2023gorillalargelanguagemodel}. The VLM's core task then becomes learning to effectively select, invoke, and interpret the results of these tools based on the visual and textual context. This approach offers modularity and scalability, allowing new capabilities to be added by simply defining new tools. However, current implementations often rely on specific prompting strategies or multi-turn conversational formats \citep{Lu2023ChameleonPC}, where the VLM generates a request for a tool call, an external system parses this request, executes the tool, and returns the result in a subsequent prompt. This imposes overhead and requires the VLM to implicitly learn the interaction protocol. A key challenge is how to design more seamless and tightly integrated tool-calling mechanisms. Could tool invocation become an intrinsic part of the VLM's generation process, perhaps through specialized tokens or architectural modifications, rather than relying on external parsing and multi-step interactions? Achieving such integration could significantly reduce latency, allow for more fluid reasoning that blends internal knowledge with external tool results, and potentially simplify the learning process for effective tool utilization. Exploring these more native tool integration strategies represents a vital step towards realizing the full potential of the tool-use paradigm for building highly capable and versatile AVLMs.

\subsection{Future Directions}

\subsubsection{Towards Unified Computational Frameworks}
Current approaches often treat external tools or knowledge sources as distinct modules "attached" to a pre-existing VLM \citep{Hu2022RevealRV}. A significant future direction involves developing more deeply integrated frameworks where the VLM and the broader computational environment (operating systems, file systems, APIs, web browsers, software applications) function as a cohesive system. Instead of relying solely on predefined tool interfaces, VLMs could learn to interact more natively with system-level functionalities, potentially generating OS commands, interacting with graphical user interface (GUI) elements directly \citep{koh2024visualwebarenaevaluatingmultimodalagents, Niu2024ScreenAgentAV}, or manipulating data structures within running applications. This paradigm shift views the AVLM not just as a language model calling tools, but as a cognitive agent situated within a digital environment, capable of complex task execution and information synthesis across diverse digital resources \citep{park2023generativeagentsinteractive}. Achieving this requires research into robust grounding of language to computational actions, secure execution environments, and models capable of long-horizon planning and interaction within complex software ecosystems.

\subsubsection{Enhancing Scalability and Efficiency via Integration}
While symbolic computations are often inherently efficient, the large neural components dominate the computational cost and parameter count in augmented systems. A key promise of neural-symbolic integration and tool use is the potential to alleviate the burden on the VLM itself. By offloading specialized tasks (such as precise calculation, factual retrieval from vast knowledge bases, complex geometric reasoning, or executing code) to external modules, the VLM may require fewer parameters dedicated to mastering these skills internally. The VLM's role shifts towards understanding the input, determining the appropriate tool or knowledge source, formulating the query or command correctly, and integrating the returned result into its ongoing reasoning process \citep{Mialon2023AugmentedLM}. This "neuro-symbolic division of labor" could lead to smaller, more efficient VLMs for certain capabilities compared to monolithic models attempting to internalize all knowledge and skills. However, the extent of this parameter reduction is an open question. While specialized skills might be effectively outsourced, the core VLM still needs substantial capacity for robust perception, language understanding, commonsense reasoning, and learning the complex skill of \textit{how and when} to interact with external systems. Research into the scaling laws governing these augmented systems is needed. Analogous to how scaling laws for pretraining relate model size, data, and compute to performance \citep{kaplan2020scalinglaws, hoffmann2022trainingcomputoptimal}, unique scaling principles might emerge for AVLMs. Factors like the number and diversity of tools, the complexity of the interaction interface, and the amount of training data demonstrating successful tool use could become critical variables influencing optimal model size and overall system performance. Understanding these "integration scaling laws" will be vital for designing compute-optimal AVLMs that effectively balance internal VLM capacity with external capabilities.

\subsubsection{Improving Generalization and Robustness with Structured Interaction}
Integrating external information introduces challenges like noisy retrieval or brittle symbolic modules, demanding robust error handling and uncertainty management \citep{Jiang2020HowCA, Gal2016DropoutAA}. Beyond mitigating errors from external sources, a crucial aspect for improving generalization and robustness lies in the nature of the interaction between the VLM and the external world. A significant advancement is enabling VLMs to produce structured outputs such as JSON objects, XML, function call arguments, SQL queries, or logical forms, instead of just natural language text \citep{Suris2023ViperGPTVI, Bhaisaheb2023ProgramSF, Gupta2022VisualPC}. This capability is fundamental for integrating AVLMs seamlessly into larger software ecosystems, which often rely on precisely formatted data exchange via APIs or databases. By generating structured data directly, AVLMs can act as reliable components within automated workflows, reducing the ambiguity and parsing errors associated with processing free-form text. Future AVLMs should increasingly be designed and trained to generate verifiable, structured representations of their reasoning or intended actions. This enhances reliability and allows for automated checks and balances within the broader system. Generalizing this concept, the future may see AVLMs interacting with computational environments through well-defined, structured protocols, enabling more complex, verifiable, and robust task execution across diverse digital and physical systems. This includes techniques for uncertainty quantification during retrieval \citep{Gal2016DropoutAA}, detecting conflicting information, adversarial training against interaction failures \citep{Wallace2019UniversalAT}, and incorporating automated verification or fact-checking \citep{Thorne2018FEVERAL}.

\subsubsection{Advancing Interpretability and Human-in-the-Loop Collaboration}
While symbolic components can inherently enhance interpretability by providing traceable reasoning steps (e.g., KG paths, rule applications, program execution logs), realizing this potential fully requires dedicated effort. Future AVLMs should be designed with interpretability as a core objective, generating not just answers but also clear, verifiable explanations grounded in the external information used \citep{Ribeiro2016WhySI}. This involves developing methods to summarize complex retrieval or reasoning processes into human-understandable narratives or visualizations. Furthermore, moving beyond passive explanation towards active human-in-the-loop systems is crucial \citep{Cai2019HumanCenteredTI}. This could involve enabling users to query the model's reasoning process, inspect intermediate results, provide feedback to correct erroneous steps, inject constraints, or guide the search for information, fostering true collaborative problem-solving between humans and AVLMs.

\subsubsection{Driving Application-Specific Advances}
The generic framework of AVLMs holds immense potential across diverse domains, but unlocking this requires tailoring integrations to specific application needs. Beyond current examples, future work should focus on developing specialized AVLMs for fields like scientific discovery (e.g., interpreting experimental data, generating hypotheses by querying scientific literature and databases \citep{Chen2020NeuralSR}), personalized education (e.g., adaptive tutoring systems that model student knowledge and retrieve relevant educational resources \citep{Aditya2019IntegratingKA}), financial analysis (e.g., systems that integrate numerical calculations, regulatory knowledge, and analysis of textual reports), and creative content generation grounded in specific world knowledge or artistic styles. This necessitates building domain-specific knowledge graphs, ontologies, symbolic solvers (e.g., physics simulators, chemical reaction predictors), and incorporating safety constraints relevant to high-stakes applications like healthcare or autonomous systems \citep{marcus2020decadeaistepsrobust}.

\section{Conclusion}

Vision-Language Models have revolutionized AI's ability to connect vision and language, yet standalone models struggle with factual accuracy, complex reasoning, adaptability, and interpretability. This systematic review charted the landscape of Augmented Vision-Language Models (AVLMs), which overcome these limitations by integrating VLMs with external symbolic information systems and computational tools. We surveyed a diverse range of techniques, categorizing them by fusion timing (early, middle, late) and the nature of augmentation (retrieval, symbolic computation, combined), revealing a clear consensus: augmenting VLMs significantly boosts performance on knowledge-intensive and reasoning-heavy tasks by synergizing neural pattern recognition with symbolic precision. A particularly powerful paradigm emerging from this landscape is tool use, which offers a flexible and unifying abstraction for AVLM design. This approach frames the VLM as an intelligent orchestrator learning to select and utilize external capabilities (such as knowledge bases, calculators, code execution, specialized algorithms, formal reasoners) encapsulated as "tools," enabling modularity and scalability. While current tool use often relies on somewhat cumbersome interaction protocols, the core concept paves the way for future systems where VLMs seamlessly integrate external resources. Significant challenges remain in managing interaction complexity, ensuring scalability and efficiency, guaranteeing robustness against unreliable external inputs, developing comprehensive evaluation methods, enhancing interpretability, and refining the tool integration mechanisms themselves. Nevertheless, the advancement of AVLMs, particularly through the lens of tool use, represents a crucial progression towards more capable, reliable, and trustworthy AI systems that effectively blend neural perception with symbolic reasoning, allowing them to not only see and describe the world but also reason about it with greater depth, accuracy, and transparency.

\newpage 
\appendix
\FloatBarrier 

\section{Methodology}
\label{methodology}
This section describes the process of gathering relevant articles for this survey, following the Preferred Reporting Items for Systematic reviews and Meta-Analyses (PRISMA) guidelines. The goal of this approach is to avoid bias when selecting what papers to review, focusing on the merits of the paper and the relevancy to the topic of AVLMs. 

\subsection{Search Strategy}

\subsubsection{Databases and Search Queries}

We utilized two primary databases for our literature search:

\begin{itemize}
    \item \textbf{Google Scholar}: Known for its extensive coverage of scholarly publications across disciplines.
    \item \textbf{Semantic Scholar}: Provides advanced search capabilities and citation analysis, facilitating the identification of semantically relevant works.
\end{itemize}

\subsubsection{Search Terms}

We formulated specific search queries to capture studies related to augmented vision-language models interacting with symbolic systems during inference. The search strategy used the strengths of both databases by employing an iterative process of testing and refining the search query until the resulting set of papers was adequately relevant. Google Scholar is more sensitive to the inclusion of keywords, and so we used a combination of Boolean operators to refine the results effectively.

The search query used in Google Scholar was:
\begin{verbatim}
"("augmented" OR "knowledge" OR "knowledge graphs" OR
"knowledge augmentation" OR "commonsense knowledge" OR
"commonsense reasoning" OR "tool use" OR
"retrieval augmented" OR "retrieval-augmented" OR
"external knowledge" OR "neural symbolic" OR
"neural-symbolic" OR "symbolic")
AND
("vision-language" OR "vision language" OR
"visual question answering" OR "image question answering" OR
"video question answering" OR "image caption" OR
"video caption" OR "image text" OR "spatial reasoning" OR
"visual reasoning")
AND
("neural network" OR "machine learning" OR
"artificial intelligence" OR "deep learning")
-"virtual reality" -"augmented reality"
\end{verbatim}

Semantic Scholar is less sensitive to keywords and more of a semantic search, so for this database, we employed a set of targeted queries to capture key aspects of our research focus:

\begin{itemize}
    \item "Commonsense reasoning in visual question answering"
    \item "Knowledge graphs for image or video captioning"
    \item "External knowledge in visual reasoning"
    \item "Neural-symbolic vision-language models"
    \item "Tool use in vision-language tasks"
    \item "Retrieval-augmented image question answering"
    \item "Symbolic reasoning in AI for vision"
    \item "Commonsense in image-text models"
    \item "Neural-symbolic visual question answering"
    \item "Multimodal knowledge graph LLM"
\end{itemize}

\subsection{Inclusion and Exclusion Criteria}

To ensure the relevance and quality of the studies included in this review, we established clear inclusion and exclusion criteria.

\subsubsection{Inclusion Criteria}

\begin{itemize}
    \item \textbf{Relevance}: Studies that describe machine learning models integrating external symbolic information systems during inference.
    \item \textbf{Language}: Publications written in English.
    \item \textbf{Implementation Focus}: Papers providing detailed descriptions of implementation methods rather than purely conceptual or theoretical discussions.
    \item \textbf{Vision-Language Tasks}: Research focusing on tasks such as visual question answering, image captioning, and video captioning where the input is imagery and/or text and the output is natural language text.
\end{itemize}

\subsubsection{Exclusion Criteria}

We excluded studies that did not align with the focus of this review, such as:

\begin{itemize}
    \item \textbf{Prompting Techniques}: Research solely on prompt engineering or techniques that rely on internal reasoning patterns without external data augmentation (e.g., chain-of-thought prompting).
    \item \textbf{Self-Prompting/Recursive Prompting}: Methods that involve iterative querying without integration of external symbolic information systems.
    \item \textbf{Synthetic Data Generation}: Studies focusing on generating synthetic data to improve model performance without external symbolic system interaction.
    \item \textbf{Architectural Modifications Without External Integration}: Papers discussing model architectures like vision encoder adapters for large language models that do not involve external symbolic systems during inference.
    \item \textbf{Training with Structured Knowledge}: Research that involves training models with external knowledge but does not allow for the external knowledge to be modified or read during inference (e.g., methods where external knowledge is embedded in model parameters).
\end{itemize}

\subsection{Selection Process}

The selection process involved several iterative steps to refine and identify the most relevant studies.

\subsubsection{Initial Search Results}
\begin{itemize}
    \item \textbf{Google Scholar}: The search yielded \textbf{980 papers} after filtering by category and removing irrelevant results based on titles and abstracts.
    \item \textbf{Semantic Scholar}: The targeted queries returned \textbf{1,332 papers}.
\end{itemize}

\subsubsection{Total Papers Collected}

In total, \textbf{2,312 papers} were collected from both databases.

\subsubsection{Relevance Scoring}

In alignment with the theme of augmented models, we utilized the \textbf{GPT-4o OpenAI model (gpt-4o-2024-08-06)} to assist in the relevance assessment:

\begin{itemize}
    \item \textbf{Automated Categorization}: GPT-4o was prompted 
    to categorize each paper and assign a relevance score ranging from 1 to 10 based on the alignment with the review topic.
    \item \textbf{Threshold for Inclusion}: Papers scoring less than \textbf{8 out of 10} were excluded from further consideration.
    \item \textbf{Iteration and Validation}: The relevance scoring process was iterated, and we ensured that all highly relevant papers were retained, even if they narrowly missed the initial threshold.
\end{itemize}

\subsubsection{Manual Screening}

\begin{figure}[htbp] 
    \centering
    \includegraphics[width=0.4\textwidth]{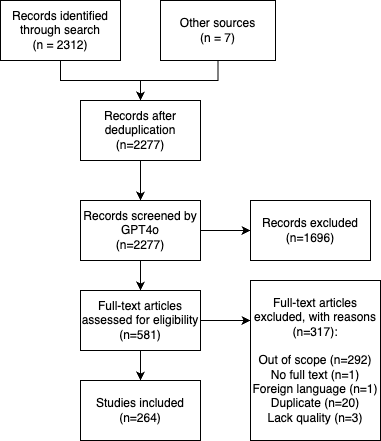} 
    \caption{PRISMA Flowchart}
    \label{fig:prisma_appendix} 
\end{figure}

\begin{itemize}
    \item \textbf{Total Papers After Automated Filtering}: \textbf{616 papers} remained after applying the relevance threshold.
    \item \textbf{Full-Text Assessment}: We conducted a thorough manual review of the full text of these papers.
    \item \textbf{Final Selection}: After removing duplicates and papers not meeting the inclusion criteria, \textbf{264 papers} were selected for detailed analysis. See Figure \ref{fig:prisma_appendix}. 
\end{itemize}

\subsection{Data Extraction and Synthesis}

From the selected studies, we extracted pertinent information to facilitate a comprehensive understanding of the methods:

\begin{itemize}
    \item \textbf{Integration Techniques}: Description of how external systems were integrated with vision-language models, classified into early fusion, middle fusion, and late fusion methods.
    \item \textbf{Types of External Symbolic Systems}: Categorization of external symbolic systems used, such as knowledge graphs, symbolic logic engines, and program synthesis tools.
    \item \textbf{Tasks Addressed}: Identification of the specific vision-language tasks tackled by each study, including visual question answering, image captioning, and others.
    \item \textbf{Implementation Details}: Detailed examination of the models' architectures, including the interaction mechanisms with external symbolic systems during inference.
\end{itemize}

\subsection{Quality Assessment}
We assessed the quality of the included studies based on:

\begin{itemize}
    \item \textbf{Clarity of Methodology}: Transparency and reproducibility of the methods described.
    \item \textbf{Experimental Rigour}: Adequacy of experimental design, including dataset usage, evaluation protocols, and statistical significance of results.
    \item \textbf{Contribution to the Field}: The extent to which the study advanced understanding or provided innovative solutions in augmented vision-language models.
\end{itemize}

\subsection{Limitations}

While we aimed for a comprehensive review, certain limitations exist:

\begin{itemize}
    \item \textbf{Publication Bias}: Unpublished works or those not indexed in the selected databases may have been missed.
    \item \textbf{Language Restriction}: Non-English publications were excluded, which may omit relevant research conducted in other languages.
    \item \textbf{Dynamic Field}: Given the rapidly evolving nature of machine learning research, new studies may have emerged after the completion of our search.
    \item \textbf{AI Bias}: The use of GPT4o in filtering of papers could potentially remove relevant search results.
\end{itemize}

By following this systematic approach, we ensured a thorough and unbiased selection of relevant literature, providing a solid foundation for the subsequent analysis and discussion in this review.

\section{Categorization Tables} 

This section contains the tables categorizing the surveyed papers based on the fusion method (Early, Middle, Late) and the type of augmentation (Retrieval, Symbolic Computation, Combined). It also includes a table summarizing relevant datasets. These tables correspond to the synthesis presented in the main Results section.

 \begin{table}[htbp]
  \centering
  \caption{Early Fusion Methods in Vision-Language Model Augmentation}
  \label{tab:early_fusion_survey_segmented_consistent_final}
  \small 

  \begin{tabular}{*{4}{c}}
    \toprule
    \multicolumn{4}{c}{\textbf{Retrieval}} \\
    \cmidrule(lr){1-4} 
    \multicolumn{2}{c}{Prompt Augmentation} & \multicolumn{1}{c}{Querying KG} & \multicolumn{1}{c}{\textbf{Retrieval Encoders}} \\
    \cmidrule(lr){1-2} \cmidrule(lr){3-3} \cmidrule(lr){4-4} 
    Image Caption & Retrieval FT & Prompt Augmentation & Subgraph Enc \\
    \midrule
    \makecell{ 
        \cite{Gao2022TransformRetrieveGenerateNL} \\ \cite{An2024KnowledgeAD} \\ \cite{Li2018VisualQA} \\ \cite{Fabian2023MultimodalFM} \\
        \cite{Sharifymoghaddam2024UniRAGUR} \\ \cite{Ghosal2023LanguageGV} \\ \cite{Khademi2023MMReasonerAM} \\ \cite{Fu2023GenerateTS} \\
        \cite{Dey2021ExternalKA} \\ \cite{Lin2022REVIVERV} \\ \cite{Yu2019MultisourceMA} \\ \cite{Liu2024CounterfactualVD} \\
        \cite{Mogadala2019MultiviewRL} \\ \cite{Lin2022RetrievalAV} \\ \cite{Garcia-Olano2021ImprovingAD} \\ \cite{Salemi2023PreTrainingMD} \\
        \cite{Luo2021WeaklySupervisedVF} \\ \cite{Vo2022NOCREKNO} \\ \cite{Hao2024SelfBootstrappedVM} \\ \cite{Gui2021KATAK} \\
        \cite{Chen2021KBVLPKB} \\ \cite{Liang2021MariaAV} \\ \cite{Lerner2023MultimodalIC}
    } &
    \makecell{ 
        \cite{Kan2023KnowledgeAwarePT} \\ \cite{Ranjit2023RetrievalAC} \\ \cite{Qu2024AlleviatingHI} \\ \cite{Liu2024RARRA} \\
        \cite{Yan2024EchoSightAV} \\ \cite{Xu2024ReverseIR} \\ \cite{Khaliq2024RAGARYF} \\ \cite{Xuan2024LEMMATL} \\
        \cite{Wen2024MultimodalRF} \\ \cite{Iscen2023RetrievalEnhancedCV} \\ \cite{Joshi2024RobustMM} \\ \cite{Chen2022MuRAGMR} \\
        \cite{Gur2021CrossModalRA} \\ \cite{Cui2024MOREMR} \\ \cite{Zhu2023MultimodalNL} \\ \cite{Hao2024KnowledgeCA}
    } &
    \makecell{ 
        \cite{Ravi2022VLCBERTVQ} \\ \cite{Narasimhan2018StraightTT} \\ \cite{Vickers2021InFE} \\ \cite{Guo2022AUE} \\
        \cite{Wang2015ExplicitKR} \\ \cite{Natu2023ExternalCK} \\ \cite{Jhalani2024PrecisionEE} \\ \cite{Barezi2024FindTG} \\
        \cite{Zhang2024GRACEGC} \\ \cite{Zhang2023KnowledgeAwareCI} \\ \cite{Wang2023DifferentiableOD} \\ \cite{Ogawa2024PredictionOA} \\
        \cite{Chen2022LaKoKV} \\ \cite{Kan2021ZeroShotSG} \\ \cite{Gan2023OpenSetKV} \\ \cite{Yang2019KnowledgeableSA}
    } &
    \makecell{ 
        \cite{Li2020BoostingVQ} \\ \cite{Rao2023RetrievalbasedKA} \\ \cite{Zhang2020RichVK} \\ \cite{Lee2021VisionLanguageKnowledgeCF} \\
        \cite{Li2022SupportingVM} \\ \cite{Lin2023TowardsMA} \\ \cite{Torino2020SemanticsAwareVA} \\ \cite{Wang2022KnowledgeEnhancedVQ} \\
        \cite{Qu2020KSFSTVC} \\ \cite{Padhi2024ImprovingCC} \\ \cite{Shevchenko2021ReasoningOV} \\ \cite{Gardres2020ConceptBertCR} \\
        \cite{Jing2023MultisourceSG} \\ \cite{Mondal2024KAMCoTKA} \\ \cite{Lee2024MultimodalRW}
    } \\
    \bottomrule
  \end{tabular}

  \medskip 

  \begin{tabular}{*{4}{c}}
    \toprule
    \multicolumn{4}{c}{\textbf{Retrieval}} \\
    \cmidrule(lr){1-4} 
    \multicolumn{4}{c}{\textbf{Retrieval Encoders (Continued)}} \\
    \cmidrule(lr){1-4} 
    \multicolumn{2}{c}{KG Encoding} & \multicolumn{2}{c}{Encoder Architectures} \\
    \cmidrule(lr){1-2} \cmidrule(lr){3-4} 
    KG Conv & MMKG Attn & Attention & LSTM \\
    \midrule
    \makecell{ 
        \cite{Chen2021ZeroshotVQ} \\ \cite{Ziaeefard2020TowardsKV} \\ \cite{Yu2020CrossmodalKR} \\ \cite{Zhu2020MuckoMC} \\
        \cite{Hussain2022MultimodalKR} \\ \cite{Li2022DynamicKM} \\ \cite{Ye2021BreakingSB}
    } &
    \makecell{ 
        \cite{Jiang2023KnowledgeBasedVQ}
    } &
    \makecell{ 
        \cite{Yuan2023RAMMRB} \\ \cite{Weng2024LearningTC} \\ \cite{Chen2022KRITKI} \\ \cite{Salemi2023ASD}
    } &
    \makecell{ 
        \cite{Wu2016ImageCA}
    } \\
    \bottomrule
  \end{tabular}

  \medskip 

  \begin{tabular}{*{2}{c}} 
    \toprule
    \multicolumn{1}{c}{\textbf{Retrieval}} & \multicolumn{1}{c}{\textbf{Symbolic}} \\
    \cmidrule(lr){1-1} \cmidrule(lr){2-2} 
    \multicolumn{1}{c}{\textbf{Retrieval Encoders (cont.)}} & \multicolumn{1}{c}{} \\ 
    \cmidrule(lr){1-1} \cmidrule(lr){2-2} 
    Memory & Symbolic \\ 
    \midrule
    \makecell{ 
        \cite{Hu2022RevealRV}
    } &
    \makecell{ 
        \cite{Potapov2019CognitiveMN}
    } \\
    \bottomrule
  \end{tabular}
\end{table}

\begin{table}[htbp]
  \centering
  \caption{Middle Fusion Methods in Vision-Language Model Augmentation}
  \label{tab:middle_fusion_survey_segmented}
  \scriptsize 

  \begin{tabular}{*{4}{c}}
    \toprule
    \multicolumn{4}{c}{\textbf{Retrieval}} \\
    \cmidrule(lr){1-4}
    \multicolumn{1}{c}{Dense Retrieval} & \multicolumn{3}{c}{Graph} \\
    \cmidrule(lr){1-1} \cmidrule(lr){2-4}
     & KG Prompt Augmentation & KG/NN Similarity & \makecell{Concept/Scene \\ Fusion} \\ 
    \midrule
    \makecell{ 
        \cite{Wang2022VisuallyAugmentedLM} \\ \cite{Lin2023FinegrainedLM} \\ \cite{Jia2023KAFARI}
    } &
    \makecell{ 
        \cite{Li2017IncorporatingEK} \\ \cite{Li2023GraphAdapterTV} \\ \cite{Zheng2021KnowledgeBG} \\ \cite{Su2018LearningVK} \\
        \cite{2018OutOT} \\ \cite{Zhang2023KnowledgeenhancedVP} \\ \cite{Singh2019FromST} \\ \cite{Jiang2020UnderstandingCI} \\
        \cite{Yu2023KnowledgeAwareGR} \\ \cite{Du2022CALMCK} \\ \cite{Li2024HKFNetFE} \\ \cite{Yin2023MultiClueRW} \\
        \cite{Ma2022MultiSourceKR} \\ \cite{Zhu2020ConfigurableGR} \\ \cite{Cao2019ExplainableHV} \\ \cite{Li2022LearningTR} \\
        \cite{ZhengSageAM} \\ \cite{Wei2022EnhanceUA} \\ \cite{Narayanan2021VQAAA}
    } &
    \makecell{ 
        \cite{Wu2024ResolvingZA} \\ \cite{Chae2022UncertaintybasedVQ} \\ \cite{Li2019SemanticCN} \\ \cite{Xian2023MultimodalKT} \\
        \cite{Marino2020KRISPII}
    } &
    \makecell{ 
        \cite{Yang2023VisualQA} \\ \cite{Wang2022VQAGNNRW} \\ \cite{Khan2022NeuSIRENI} \\ \cite{Khan2022ExpressiveSG} \\
        \cite{Zhu2022FromST} \\ \cite{Wen2021MultiLevelKI} \\ \cite{Song2023HowTU} \\ \cite{Li2022IncorporatingEK} \\
        \cite{Zhang2021ExplicitKI} \\ \cite{Dong2024ModalityAwareIW} \\ \cite{Gao2023RoomObjectEP} \\ \cite{Zhang2022ReasoningWM} \\
        \cite{Xu2021ImagineRA} \\ \cite{Li2024IIUII} \\ \cite{Hou2020JointCA} \\ \cite{Gu2019SceneGG}
    } \\
    \bottomrule
  \end{tabular}

  \medskip 

  \begin{tabular}{*{4}{c}}
    \toprule
    \multicolumn{3}{c}{\textbf{Retrieval}} & \multicolumn{1}{c}{\textbf{Symbolic Computation}} \\
    \cmidrule(lr){1-3} \cmidrule(lr){4-4}
    \multicolumn{2}{c}{Graph} & \multicolumn{1}{c}{RL} & \multicolumn{1}{c}{Program Synthesis} \\
    \cmidrule(lr){1-2} \cmidrule(lr){3-3} \cmidrule(lr){4-4}
    MMKGs & Hypergraphs &  &  \\
    \midrule
     \makecell{ 
        \cite{Xi2024KnowledgeGE} \\ \cite{Shi2022ImprovingZP} \\ \cite{Santiesteban2024ImprovedIC} \\ \cite{Ouyang2024ModaladaptiveKG} \\
        \cite{Liu2021FactbasedVQ}
    } &
    \makecell{ 
        \cite{Heo2022HypergraphTW} \\ \cite{Wang2024HyperMRHH}
    } &
    \makecell{ 
        \cite{Bougie2018CombiningDR}
    } &
    \makecell{ 
        \cite{Zhang2022QueryAA} \\ \cite{Zhang2023TowardMD} \\ \cite{Hu2023VisualPD} \\ \cite{Shirai2023VisionLanguageIF} \\
        \cite{Zhang2023GroundingCT} \\ \cite{Li2021CalibratingCA} \\ \cite{Mishra2024LearningRF} \\ \cite{Xue2024IntegratingNR}
    } \\
    \bottomrule
  \end{tabular}

  \medskip 

  \begin{tabular}{*{4}{c}}
    \toprule
    \multicolumn{4}{c}{\textbf{Symbolic Computation}} \\
    \cmidrule(lr){1-4}
    Logic Engines & VSA & \makecell{Symbolic Graph Ops} & Tool Use \\ 
    \midrule
    \makecell{ 
        \cite{Amizadeh2020NeuroSymbolicVR} \\ \cite{Riley2019IntegratingNL} \\ \cite{Mitchener2021DetectUA} \\ \cite{Tsatsou2021TowardsUK} \\
        \cite{Zhang2022DANLIDA} \\ \cite{Choi2024TowardsNV} \\ \cite{Li2023AND} \\ \cite{Li2023ScallopAL} \\
        \cite{Huang2021ScallopFP} \\ \cite{Zhang2023SEAGULLAE}
    } &
    \makecell{ 
        \cite{Montone2017HyperdimensionalCF} \\ \cite{Kovalev2021VectorSM}
    } &
    \makecell{ 
        \cite{Li2022InnerKI} \\ \cite{Liang2020LRTAAT} \\ \cite{Wu2023SymbolLLMLL} \\ \cite{Zhao2015AQF} \\
        \cite{Yang2020ObjectCentricDO} \\ \cite{Zhang2023InterpretableDO} \\ \cite{Hudson2019LearningBA} \\ \cite{Cao2021LinguisticallyRC}
    } &
    \makecell{ 
        \cite{Hu2024VisualSS} \\ \cite{Fan2024VideoAgentAM} \\ \cite{Liu2023LLaVAPlusLT} \\ \cite{Hu2023AVISAV} \\
        \cite{Wu2024AvaTaROL}
    } \\
    \bottomrule
  \end{tabular}

  \medskip 

  \begin{tabular}{*{3}{c}} 
    \toprule
    \multicolumn{2}{c}{\textbf{Symbolic Computation}} & \multicolumn{1}{c}{\textbf{Combined Retr \& Symb}} \\
    \cmidrule(lr){1-2} \cmidrule(lr){3-3}
    Self Play & Agents & Other \\
    \midrule
    \makecell{ 
        \cite{Misiunas2024VQATS}
    } &
    \makecell{ 
        \cite{Niu2024ScreenAgentAV} \\ \cite{Castrejon2024HAMMRHM} \\ \cite{Lu2023ChameleonPC} \\ \cite{Hsieh2023ToolDE} \\
        \cite{Xu2024RSAgentAR} \\ \cite{Yang2024DoraemonGPTTU}
    } &
    \makecell{ 
        \cite{Besbes2015AnOV} \\ \cite{Aditya2016AnsweringIR} \\ \cite{Aditya2017ExplainableIU} \\ \cite{Aditya2016DeepIUA} \\
        \cite{Tan2021KnowledgeBasedEQ} \\ \cite{Liu2023KnowledgeEnhancedHI} \\ \cite{Stammer2024NeuralCB} \\ \cite{Vatashsky2018UnderstandCA} \\
        \cite{Gao2023ASR} \\ \cite{Gao2024ExplainableVQ}
    } \\
    \bottomrule
  \end{tabular}
\end{table}

\begin{table}[htbp]
  \centering
  \caption{Late Fusion Methods in Vision-Language Model Augmentation}
  \label{tab:late_fusion_survey_segmented}
  \small 

  \begin{tabular}{*{4}{c}}
    \toprule
    \multicolumn{2}{c}{\textbf{Retrieval}} & \multicolumn{2}{c}{\textbf{Symbolic Computation}} \\
    \cmidrule(lr){1-2} \cmidrule(lr){3-4}
    Dense & Knowledge Graph & Program Synth & Symbolic Engines \\
    \midrule
    \makecell{ 
        \cite{Song2022AnsweringKV} \\ \cite{Song2022EfficientAS} \\ \cite{Shi2024QRCLIPIE}
    } &
    \makecell{ 
        \cite{Gao2022CrossModalOD} \\ \cite{Huang2020BoostIC} \\ \cite{Xiao2022VisualRD}
    } &
    \makecell{ 
        \cite{Vedantam2019ProbabilisticNM} \\ \cite{Yi2018NeuralSymbolicVD} \\ \cite{Suris2023ViperGPTVI} \\ \cite{Khandelwal2023AnalyzingMA} \\
        \cite{Subramanian2023ModularVQ} \\ \cite{Gupta2022VisualPC} \\ \cite{Bhaisaheb2023ProgramSF}
    } &
    \makecell{ 
        \cite{Sethuraman2021VisualQA} \\ \cite{Aditya2018SpatialKD} \\ \cite{Eiter2022ANA} \\ \cite{Eiter2021ACI} \\
        \cite{Cunnington2024TheRO}
    } \\
    \bottomrule
  \end{tabular}

  \medskip 

  \begin{tabular}{*{4}{c}}
    \toprule
    \multicolumn{3}{c}{\textbf{Symbolic Computation}} & \multicolumn{1}{c}{\textbf{Combined}} \\
    \cmidrule(lr){1-3} \cmidrule(lr){4-4}
    \makecell{Symbolic Graph Ops} & Tool Use & Other & Combined \\ 
    \midrule
    \makecell{ 
        \cite{Li2023MultimodalAC} \\ \cite{Zhan2021VisualQA} \\ \cite{Saqur2020MultimodalGN} \\ \cite{Johnston2023NSILNV}
    } &
    \makecell{ 
        \cite{Yuan2023CRAFTCL} \\ \cite{Cesista2024RetrievalAS} \\ \cite{Cesista2024MultimodalSG} \\ \cite{Zhang2023GraphToolFormerTE}
    } &
    \makecell{ 
        \cite{Xu2022SGLSG} \\ \cite{Singh2018ABA} \\ \cite{Bao2023ConfidencebasedIN} \\ \cite{Verheyen2023NeuroSymbolicPS}
    } &
    \makecell{ 
        \cite{Sachan2020TowardsLA} \\ \cite{Basu2020AQuAAV}
    } \\
    \bottomrule
  \end{tabular}
\end{table}

\begin{table}[htbp]
  \centering
  \caption{Datasets Relevant to Augmented Vision-Language Models}
  \label{tab:datasets_survey_segmented}
  \small 

  \begin{tabular}{*{4}{c}}
    \toprule
    \multicolumn{2}{c}{\textbf{Spatial Reasoning}} & \multicolumn{1}{c}{\textbf{Knowledge Based VQA}} & \multicolumn{1}{c}{\textbf{Reasoning VQA}} \\
    \cmidrule(lr){1-2} \cmidrule(lr){3-3} \cmidrule(lr){4-4}
    CLEVER & Scene Graph & KBVQA & Reasoning VQA \\
    \midrule
    \makecell{ 
        \cite{johnson2016clevrdiagnosticdatasetcompositional} \\ \cite{yi2020clevrercollisioneventsvideo} \\ \cite{li2023superclevrvirtualbenchmarkdiagnose} \\ \cite{Abraham2024CLEVRPOCRV} \\ \cite{Wang2024Compositional4D}
    } &
    \makecell{ 
        \cite{Krishna2016VisualGC} \\ \cite{Shen2024VCDKB}
    } &
    \makecell{ 
        \cite{agrawal2016vqavisualquestionanswering} \\ \cite{Wang2016FVQAFV} \\ \cite{Lin2023FVQA2I} \\ \cite{Shah2019KVQAKV} \\
        \cite{Marino2019OKVQAAV} \\ \cite{Reichman2023OutsideKV} \\ \cite{Su2024SKVQASK} \\ \cite{Mensink2023EncyclopedicVV} \\
        \cite{Jain2021SelectSS} \\ \cite{Cao2020KnowledgeRoutedVQ} \\ \cite{Sung2022INKI} \\ \cite{Agarwal2024IndiFoodVQAAV} \\
        \cite{Qiu2024SnapNTellEE} \\ \cite{Lerner2022ViQuAEAD}
    } &
    \makecell{ 
        \cite{Schwenk2022AOKVQAAB} \\ \cite{Gao2019CRICAV} \\ \cite{Zellers2018FromRT} \\ \cite{Cao2019ExplainableHV} \\
        \cite{BittonGuetta2024VisualRA}
    } \\
    \bottomrule
  \end{tabular}

  \medskip 

  \begin{tabular}{*{4}{c}}
    \toprule
    \multicolumn{1}{c}{\textbf{Knowledge and Spatial}} & \multicolumn{1}{c}{\textbf{Agents}} & \multicolumn{2}{c}{\textbf{Task Specific}} \\
    \cmidrule(lr){1-1} \cmidrule(lr){2-2} \cmidrule(lr){3-4}
    Knowledge and Spatial & Agents & Robotics & \makecell{Other (Task)} \\ 
    \midrule
    \makecell{ 
        \cite{Chen2023CanPV} \\ \cite{Wang2024SOKBenchAS} \\ \cite{Zhang2023CanVM}
    } &
    \makecell{ 
        \cite{Niu2024ScreenAgentAV} \\ \cite{zhou2023webarena} \\ \cite{Cao2024Spider2VHF}
    } &
    \makecell{ 
        \cite{Gao2023PhysicallyGV}
    } &
    \makecell{ 
        \cite{Hayashi2024TowardsAE} \\ \cite{Jin2024FakeND} \\ \cite{Hu2023ExpertKI}
    } \\
    \bottomrule
  \end{tabular}
\end{table}

\FloatBarrier 

\bibliography{bibs} 
\bibliographystyle{tmlr}

\end{document}